\documentclass[runningheads]{llncs}

\usepackage{eccv}

\usepackage{graphicx}
\usepackage{booktabs}
\usepackage{multirow}
\usepackage{xcolor}
\definecolor{tabgreen}{RGB}{30, 160, 30}
\usepackage{makecell}
\usepackage{eccvabbrv}

\usepackage{graphicx}
\usepackage{booktabs}

\usepackage{amsmath}
\usepackage{amssymb}
\usepackage{mathtools}
\usepackage{amsthm}
\usepackage{amsfonts,mathrsfs}

\usepackage[accsupp]{axessibility}  % Improves PDF readability for those with disabilities.

\newcommand\R{\mathbb{R}}

\usepackage{hyperref}

\begin{document}

\title{Weight Conditioning for Smooth Optimization of Neural Networks} 


\author{Hemanth Saratchandran\inst{1} \and
Thomas X Wang\inst{2} \and
Simon Lucey\inst{1}}

\authorrunning{H.~Saratchandran et al.}

\institute{Australian Institute for Machine Learning, University of Adelaide \and 
Sorbonne Universit{\'e}, CNRS, ISIR \\
\email{hemanth.saratchandran@adelaide.edu.au}}

%\institute{Princeton University, Princeton NJ 08544, USA \and
%Springer Heidelberg, Tiergartenstr.~17, 69121 Heidelberg, Germany

\maketitle

\begin{abstract}
  In this article, we introduce a novel normalization technique for neural network weight matrices, which we term weight conditioning. This approach aims to narrow the gap between the smallest and largest singular values of the weight matrices, resulting in better-conditioned matrices. The inspiration for this technique partially derives from numerical linear algebra, where well-conditioned matrices are known to facilitate stronger convergence results for iterative solvers. We provide a theoretical foundation demonstrating that our normalization technique smoothens the loss landscape, thereby enhancing convergence of stochastic gradient descent algorithms. Empirically, we validate our normalization across various neural network architectures, including Convolutional Neural Networks (CNNs), Vision Transformers (ViT), Neural Radiance Fields (NeRF), and 3D shape modeling. Our findings indicate that our normalization method is not only competitive but also outperforms existing weight normalization techniques from the literature.
\end{abstract}

\section{Introduction} \label{sec:intro}

Normalization techniques, including batch normalization \cite{ioffe2015batch}, weight standardization \cite{qiao2019micro}, and weight normalization \cite{salimans2016weight}, have become fundamental to the advancement of deep learning, playing a critical role in the development and performance optimization of deep learning models for vision applications \cite{he2016deep, huang2023normalization, zhang2020exemplar, lubana2021beyond}. By ensuring consistent scales of inputs and internal activations, these methods not only stabilize and accelerate the convergence process but also mitigate issues such as vanishing or exploding gradients. 
In this paper, we put forth a normalization method termed \textit{weight conditioning}, designed to help in the optimization of neural network architectures, including both feedforward and convolutional layers, through strategic manipulation of their weight matrices. 
By multiplying a weight matrix of a neural architecture by a predetermined matrix conditioner, weight conditioning aims to minimize the condition number of these weight matrices, effectively narrowing the disparity between their smallest and largest singular values. Our findings demonstrate that such conditioning not only influences the Hessian matrix of the associated loss function when training such networks, leading to a lower condition number but also significantly enhances the efficiency of iterative optimization methods like gradient descent by fostering a smoother loss landscape.  
Our theoretical analysis elucidates the impact of weight matrices on the Hessian's condition number, revealing our central insight: optimizing the condition number of weight matrices directly facilitates Hessian conditioning, thereby expediting convergence in gradient-based optimization by allowing a larger learning rate to be used. Furthermore, weight conditioning can be used as a drop-in component along with the above normalization techniques, yielding better accuracy and training on a variety of problems.

\begin{figure}[t]
    \centering
    \includegraphics[width=1.0\linewidth]
    {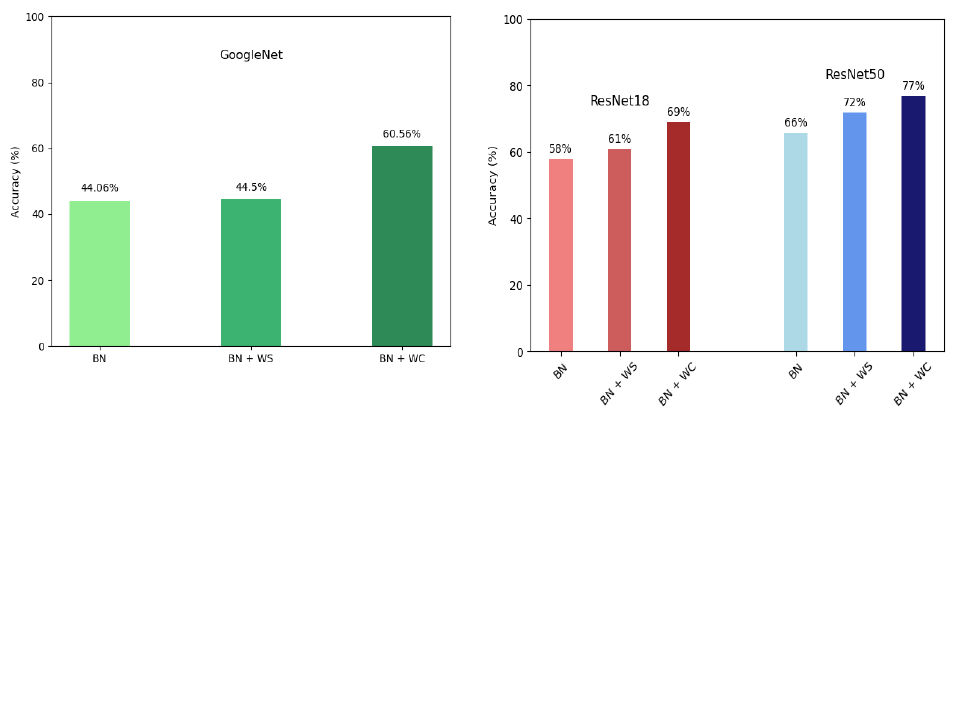}
    %\caption{Trained for 350s}
    %\end{subfigure}
    \vspace{-13em}
    \caption{Left; Testing three different normalizations on the GoogleNet CNN trained on CIFAR100. BN + WC (ours) reaches a much higher accuracy than the other two. Right; Testing the same three normalizations on a ResNet18 and ResNet50 CNN architecture. In both cases BN + WC (ours) performs better.}
    \label{fig:front_fig}
\end{figure}

We draw the reader's focus to fig. \ref{fig:front_fig}, which delineates the pivotal role of weight conditioning in enhancing batch normalization's effectiveness. Illustrated on the left, the figure compares training outcomes for a GoogleNet model on the CIFAR100 dataset under three distinct conditions: Batch Normalization (BN), Batch Normalization with Weight Standardization (BN + WS), and Batch Normalization with Weight Conditioning (BN + WC). Each variant was subjected to a training regimen of 40 epochs using Stochastic Gradient Descent (SGD) at an elevated learning rate of 2. Remarkably, the BN + WC configuration demonstrates a superior accuracy enhancement of nearly 15\% over its counterparts, showcasing its robustness and superior adaptability to higher learning rates than typically documented in the literature. On the figure's right, we extend this comparative analysis to include ResNet18 and ResNet50 models, trained on CIFAR100, via SGD with a conventional learning rate of 1e-3 and over a span of 200 epochs. Consistently, BN + WC exhibits a pronounced performance advantage over the alternative normalization strategies, reinforcing its efficacy across diverse neural network frameworks.

To further demonstrate the versatility of weight conditioning, we rigorously tested its efficacy across a spectrum of machine learning architectures, including Convolutional Neural Networks (CNNs), Vision Transformers (ViTs), Neural Radiance Fields (NeRF), and 3D shape modeling. In each scenario, we juxtaposed weight conditioning against established normalization methods cited in current research, unveiling its potential to significantly boost the performance of these advanced deep learning frameworks. 
%We found that for CNN architectures weight conditioning was competitive with weight standardization and on ViTs could be used as a drop-in component along with layer normalization to boost performance. 
Our main contributions are:
\begin{enumerate}
    \item We introduce a novel normalization strategy, termed \textit{weight conditioning}, which strategically modifies the weight matrices within neural networks, facilitating faster convergence for gradient based optimizers. 
    \item Through rigorous theoretical analysis, we validate the underlying principles of weight conditioning, offering a solid foundation for its implementation and understanding.
    \item We present comprehensive empirical evidence showcasing the effectiveness of weight conditioning across a variety of machine learning models, highlighting its broad applicability and impact on model performance optimization.
\end{enumerate}

\section{Related Work\label{sec;rel_work}}

\subsubsection{Normalization in deep learning:} Normalization techniques have become pivotal in enhancing the training stability and performance of deep learning models. Batch normalization, introduced by Ioffe and Szegedy \cite{ioffe2015batch}, normalizes the inputs across the batch to reduce internal covariate shift, significantly improving the training speed and stability of neural networks. Layer normalization, proposed by Ba et al. \cite{ba2016layer}, extends this idea by normalizing inputs across features for each sample, proving particularly effective in recurrent neural network architectures \cite{ba2016layer} and transformer networks \cite{xiong2020layer}. Weight normalization, by Salimans and Kingma \cite{salimans2016weight}, decouples the magnitude of the weights from their direction, facilitating a smoother optimization landscape. Lastly, weight standardization, introduced by Qiao et al. \cite{qiao2019micro}, standardizes the weights in convolutional layers, further aiding in the optimization process, especially when combined with batch normalization. Together, these techniques address various challenges in training deep learning models, underscoring the continuous evolution of strategies to improve model convergence and performance.

\section{Notation}\label{sec;Notation}
 Our main theoretical contributions will be in the context of feedfoward layers. Therefore, we fix notation for this here.
Let $F$ denote a depth $L$ neural network with layer widths 
$\{n_1,\ldots,n_L\}$. We let $X \in \R^{N\times n_0}$ denote the training data, with $n_0$ being the dimension of the input. The output at layer $k$ will be denoted by $F_k$ and is defined by
\begin{equation}\label{defn_net}
    F_k =  
    \begin{cases}
        F_{L-1}W_L + b_L, & k = L \\
        \phi(F_{k-1}W_k + b_k), & k \in [L-1] \\
        X, & k = 0
    \end{cases}
\end{equation}
where the weights $W_k \in \R^{n_{k-1}\times n_k}$ and the biases 
$b_k \in \R^{n_k}$ and $\phi$ is an activation applied component wise.
The notation
$[m]$ is defined by $[m] = \{1,\ldots,m\}$.
We will also fix a loss function $\mathcal{L}$ for minimizing the weights of $F$. In the experiments this will always be the MSE loss or the Binary Cross Entropy (BCE) loss. Note that $\mathcal{L}$ depends on $F$.

\section{Motivation}\label{sec;motivation}
In this section, we give some brief motivation for the theoretical framework we develop in the next section.
\paragraph{\textbf{A simple model:}}Consider the quadratic objective function given by
\begin{equation}\label{eqn;simple_model}
    \mathcal{L}(\theta) := \frac{1}{2}\theta^TA\theta - b^T\theta
\end{equation}
where $A$ is a symmetric $n \times n$ matrix of full rank and $b$ is an $n \times 1$ vector. The objective function $\mathcal{L}$ is used when solving the eqn. $Ax = b$. One can see that the solution of this equation if given by $x = A^{-1}b$ which is precisely the minimum $\theta^*$ of $\mathcal{L}$. Thus minimizing the objective function $\mathcal{L}$ with a gradient descent algorithm is one way to find a solution to the matrix equation $Ax = b$. We want to consider the gradient descent algorithm on $\mathcal{L}$ and understand how the convergence of such an algorithm depends on characteristics of the matrix $A$.

Observe that 
\begin{equation}\label{eqn;grad_hess_simple_obj}
    \nabla \mathcal{L}(\theta) = A\theta - b \text{ and } 
    H(\mathcal{L})(\theta) = A
\end{equation}
where $H(\mathcal{L})(\theta)$ denotes the Hessian of $\mathcal{L}$ at $\theta$. If we consider the gradient descent update for this objective function with a learning rate of $\eta$, eqn. \eqref{eqn;grad_hess_simple_obj} implies
\begin{equation}\label{eqn;grad_descent_up}
    \theta^{t+1} = \theta^t - \eta\nabla \mathcal{L}(\theta^t) = 
    \theta^t - \eta (A\theta^t - b)
\end{equation}
Taking the singular value decomposition (SVD) of $A$ we can write 
\begin{equation}\label{eqn;svd}
    A = Udiag(\sigma_1,\cdots ,\sigma_n)V^T
\end{equation}
where $U$ and $V$ are unitary matrices and $\sigma_1 \geq \cdots \geq \sigma_n$ are the singular values of $A$. The importance of the SVD comes from the fact that we can view the gradient descent update \eqref{eqn;grad_descent_up} in terms of the basis defined by $V^T$. Namely, we can perform a change of coordinates and define
\begin{equation}\label{eqn;change_coords1}
    x^t = V^T(\theta^t - \theta^*)
\end{equation}
The gradient update for the ith-coordinate of $x^t$, denoted $x_i^t$ becomes 
\begin{equation}\label{eqn;grad_up_change_coords}
    x^{(t+1)}_i = x^t_i - \eta\sigma_ix_i^t = (1- \eta\sigma_i)x_i^t
    = (1-\eta\sigma_i)^{t+1}x_0.
\end{equation}
If we write $V = [v_1,\cdots ,v_n]$  with each $v_i \in \R^{n\times 1}$ we then have
\begin{equation}\label{eqn;grad_up_back_coords}
    \theta^{t} - \theta^* = Vx^t = \sum_{i=1}^nx_i^0(1-\eta\sigma_i)^{t+1}
    v_i.
\end{equation}
Eqn. \eqref{eqn;grad_up_back_coords} shows that the rate at which gradient descent moves depends on the quantities 
$1-\eta\sigma_i$. This implies that in the direction $v_i$, gradient descent moves at a rate of $(1-\eta\sigma_i)^t$ from which it follows that the closer $1-\eta\sigma_i$ is to zero the faster the convergence. In particular, provided $\eta$ is small enough, the directions corresponding to the larger singular values will converge fastest.
Furthermore, eqn. \eqref{eqn;grad_up_back_coords} gives a condition on how the singular values of $A$ affect the choice of learning rate $\eta$. We see that in order to guarantee convergence we need 
\begin{equation}\label{eqn;learning_rate_bound1}
\vert 1 - \eta \sigma_i\vert < 1 \text{ for all } 1 \leq i \leq n
\end{equation}
which implies
\begin{equation}\label{eqn;rel_lr_sings}
0 < \eta\sigma_i < 2 \text{ for all } 1 \leq i \leq n.
\end{equation}
This means we must have $\eta < \frac{2}{\sigma_i}$ for each $i$ and this will be satisfied if $\eta < \frac{2}{\sigma_1}$ since $\sigma_1$ is the largest singular value. Furthermore, we see that the progress of gradient descent in the $ith$ direction $v_i$ is bounded by
\begin{equation}\label{eqn;}
    \eta\sigma_i < \frac{2\sigma_i}{\sigma_1} < 
    \frac{2\sigma_1}{\sigma_n}.
\end{equation}
Since $\sigma_1 \geq \sigma_n$ we thus see that gradient descent will converge faster provided 
the quantity $\frac{\sigma_1}{\sigma_n}$ is as small as possible.
This motivates the following definition.
\begin{definition}\label{defn;condition_num}
    Let $A$ be a $n \times m$ matrix of full rank. The condition number of $A$ is defined by 
    \begin{equation}\label{eqn;cond_A}
        \kappa(A) := \frac{\sigma_1(A)}{\sigma_k(A)}
    \end{equation}
   where $\sigma_1(A) \geq \cdots \geq \sigma_k(A) > 0$ and 
   $k = \min\{m, n\}$.
\end{definition}
Note that because we are assuming $A$ to be full rank, the condition number is well defined as all the singular values are positive.

%The above discussion shows how the rate of convergence of gradient descent for the objective function $\mathcal{L}$ is related to the condition number of $A$ and more importantly that the closer the condition number is to $1$ the faster the algorithm converges. 
%Thus a natural question that arises is how we can bring down the condition number of $A$? This naturally motivates the concept of preconditioning. 

\paragraph{\textbf{Preconditioning:}} Preconditioning involves the application of a matrix, known as the preconditioner $P$, to another matrix $A$, resulting in the product $PA$, with the aim of achieving $\kappa(PA) \leq \kappa(A)$ \cite{nocedal1999numerical, qu2022optimal}. This process, typically referred to as left preconditioning due to the multiplication of $A$ from the left by $P$, is an effective method to reduce the condition number of $A$. Besides left preconditioning, there is also right preconditioning, which considers the product $AP$, and double preconditioning, which employs two preconditioners, $P_1$ and $P_2$, to form $P_1AP_2$. Diagonal matrices are frequently chosen as preconditioners because their application involves scaling the rows or columns of $A$, thus minimally adding to the computational cost of the problem. Examples of preconditioners are:
\begin{itemize}
    \item[1.] \textbf{Jacobi Preconditioner:} Given a square matrix $A$ the Jacobi preconditioner $D$ consists of the inverse of the diagonal elements of $A$, $A \rightarrow diag(A)^{-1}A$ \cite{jacobi1845ueber}.

    \item[2.] \textbf{Row Equilibration:} Given a $n \times m$ matrix $A$, row equilibration is a diagonal $n\times n$ matrix with the inverse of the 2-norm of each row of $A$ on the diagonal, 
    $A \rightarrow (\vert\vert A_{i:}\vert\vert_2)^{-1}A$, where $A_{i:}$ denotes the ith-row of $A$ \cite{bradley2010algorithms}.

    \item[3.] \textbf{Column Equilibration:} Given a $n \times m$ matrix $A$, column equilibration is a diagonal $m\times m$ matrix with the inverse of the 2-norm of each column of $A$ on the diagonal, 
    $A \rightarrow A(\vert\vert A_{:i}\vert\vert_2)^{-1}$, where $A_{:i}$ denotes the ith-column of $A$.
    \item[4.] \textbf{Row-Column Equilibration:} This is a double sided equilibration given by row equilibration on the left and column equilibration on the right. 
\end{itemize}
The interested reader can consult \cite{nocedal1999numerical, qu2022optimal} for more on preconditioner. In Sec. \ref{subsec;condition_ff} we will explain how row equilibration helps reduce the condition number. 

If we precondition $A$ yielding $PA$ and consider the new objective function 
\begin{equation}\label{eqn;precond_simple_obj}
    \mathcal{L}_P(\theta) = \theta^TPA\theta - b^TP^T\theta
\end{equation}

Then provided we have that $\kappa(PA) \leq \kappa(A)$ the above discussion shows that gradient descent on the new objective function
$\mathcal{L}_P$ will converge faster. For this problem the preconditioning can also be thought of in terms of the matrix eqn. $Ax = b$. We seek to multiply the system by $P$ yielding the new system 
$PAx = Pb$ and provided $\kappa(PA) \leq \kappa(A)$, this new system will be easier to solve using a gradient descent.

\paragraph{\textbf{General objective functions:}} 
The above discussion focused on a very simple objective function given by eqn. \eqref{eqn;simple_model}. In general, objective functions are rarely this simple. However, given a general objective function $\widetilde{\mathcal{L}}$, about a point $\theta_0$ we can approximate $\widetilde{\mathcal{L}}$ by a second order Taylor series yielding 
\begin{equation}
    \widetilde{\mathcal{L}}(\theta) \approx 
    \frac{1}{2}(\theta - \theta_0)^TH(\theta_0)(\theta - \theta_0) + 
    (\nabla\widetilde{\mathcal{L}}(\theta_0))(\theta - \theta_0) + 
    \widetilde{\mathcal{L}}(\theta_0)
\end{equation}
where $H$, the Hessian matrix of $\widetilde{\mathcal{L}}$, is a symmetric square matrix that captures the curvature of $\widetilde{\mathcal{L}}$. This expansion offers insights into the local behavior of gradient descent around $\theta_0$, particularly illustrating that if $H$ is full rank, the convergence speed of the descent algorithm is influenced by the condition number of $H$. However, for many neural network architectures, which typically have a vast number of parameters, directly computing $H$ is impractical due to its $\mathcal{O}(n^2)$ computational complexity, where $n$ represents the parameter count. To address this challenge, the subsequent section will introduce the concept of weight conditioning, a technique aimed at reducing the condition number of $H$ without the necessity of direct computation.

\section{Theoretical Methodology}\label{sec;theory_methods}

In this section, we explain our approach to conditioning the weight matrices within neural networks, aiming to enhance the convergence efficiency of gradient descent algorithms. Fixing a loss function $\mathcal{L}$ we saw in the previous sec. \ref{sec;motivation} that we can approximate $\mathcal{L}$ locally by a second order quadratic approximation via the Taylor series about any point $\theta_0$ as
\begin{equation}\label{eqn;quad_hessian_restated}
    \mathcal{L}(\theta) \approx 
    \frac{1}{2}(\theta - \theta_0)^TH(\theta_0)(\theta - \theta_0) + 
    (\nabla\mathcal{L}(\theta_0))(\theta - \theta_0) + 
    \mathcal{L}(\theta_0)
\end{equation}
where $H$ is the Hessian matrix associated to $\mathcal{L}$. Note that $H$ is a square symmetric matrix. The discussion in sec. \ref{sec;motivation} made two key observations:
\begin{enumerate}
      \item The convergence rate of gradient descent on $\mathcal{L}$ is significantly influenced by the condition number of $H$, provided $H$ is full rank.
    \item Inspired by the first point, the convergence rate can be accelerated by diminishing the condition number of $H$.
\end{enumerate}
It was mentioned at the end of sec. \ref{sec;motivation} that in general we cannot expect to have direct access to the Hessian $H$ as this would require a computational cost of $\mathcal{O}(n^2)$, where $n$ is the number of parameters, which can be extremely large for many deep learning models. In this section we introduce weight conditioning, which is a method of conditioning the weights of a neural networks that leads to a method of bringing down the condition number of $H$ without directly accessing it.

\subsection{Weight conditioning for feed forward networks}\label{subsec;condition_ff}

\begin{figure}[t]
    \centering
    \includegraphics[width=1.0\linewidth]
    {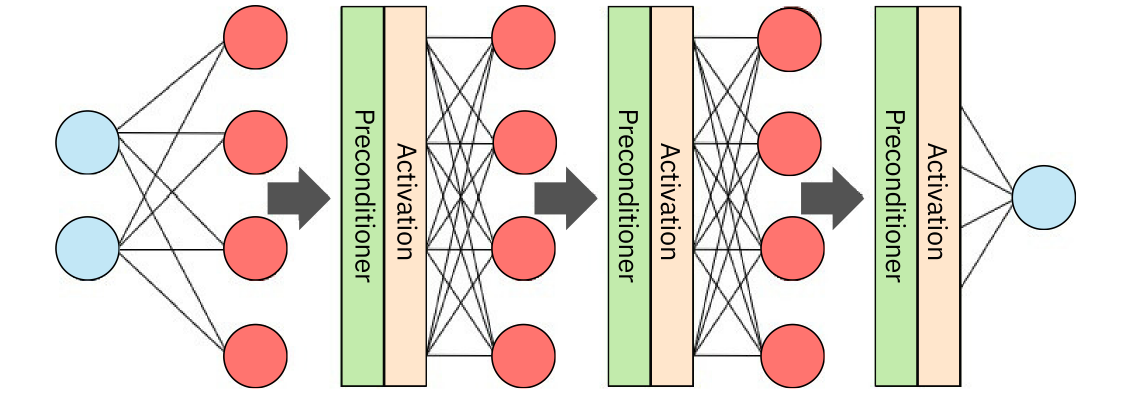}
    %\caption{Trained for 350s}
    %\end{subfigure}
  %  \vspace{-8em}
\caption{Schematic representation of a preconditioned network. The weights from the neurons (red) are first multiplied by a preconditioner matrix (green) before being activated (orange).}
    \label{fig:precond_net}
\end{figure}

We fix a neural network $F(x;\theta)$ with $L$ layers, as defined in sec. \ref{sec;Notation}. Given a collection of preconditioner matrices $P = \{P_1,\cdots, P_l\}$ where 
$P_k \in \R^{n_{k-1}\times n_k}$ we define a preconditioned network 
$F^{\textit{pre}}(x;\theta)$ as follows. The layer maps of $F^{\textit{pre}}(x;\theta)$ will be defined by:
\begin{equation}\label{eqn;precond_network_defn}
    F^{\textit{pre}}_k =
    \begin{cases}
    \phi_k((P                                     _kW_k)^T(F^{\textit{pre}}_{k-1})(x) + b_k), & k = [1, L] \\
    x, & k = 0.
    \end{cases}
\end{equation}
We thus see that the layer-wise weights of the network $F^{\textit{pre}}$ are the weights $W_k$ of the network $F$ preconditioned by the preconditioner matrix $P_k$. Fig. \ref{fig:precond_net} provides a visual depiction of the preconditioned network $F^{\textit{pre}}(x;\theta)$, illustrating how the preconditioner matrices are applied to each layer's weights to form the updated network configuration. 

\begin{definition}\label{defn;weight_cond}
    Given a feed forward neural network $F(x;\theta)$ we call the process of going from $F(x;\theta)$ to $F^{\textit{pre}}(x;\theta)$ weight conditioning. 
\end{definition}
It's important to observe that both networks, $F$ and $F^{\textit{pre}}$, maintain an identical count of parameters, activations, and layers. The sole distinction between them lies in the configuration of their weight matrices.

Our objective is to rigorously demonstrate that an optimal choice for weight conditioning a network is to use row equilibration.
Recall from sec. \ref{sec;motivation}, that row equilibration preconditioner for a weight matrix $W_k \in \mathbb{R}^{n_{k-1}\times n_k}$ is defined as a diagonal matrix $E_k \in \mathbb{R}^{n_{k-1}\times n_{k-1}}$. Each diagonal element of $E_k$ is determined by the inverse of the 2-norm of the $i$-th row vector of $W_k$, i.e. 
$\vert\vert (W_k)_{i:}\vert\vert_2^{-1}$. 

%The left hand side of fig. \ref{fig:equib} gives a schematic viewpoint of row equilibration for a matrix.

All our statements in this section hold for the column equilibrated preconditioner and the row-column equilibrated preconditioner. Therefore, from here on in we will drop the usage of the word row and simply call our preconditioner an equilibrated preconditioner.

%\begin{figure}[t]
%    \centering
%    \includegraphics[width=12cm, height=3.5cm]
%    {figs/theory/equib_final}
    %\caption{Trained for 350s}
    %\end{subfigure}
  %  \vspace{-8em}
%    \caption{sad}
%    \label{fig:equib}
%\end{figure}

There are two main reasons we are choosing to use equilibration as the preferred form to condition the weights of the neural network $F$. The first reason is that it is a diagonal preconditioner, therefore requiring low computational cost to compute. Though more importantly, as the following theorem of Van Der Sluis \cite{van1969condition} shows, it is the optimum preconditioner amongst all diagonal preconditioners to reduce the condition number.

\begin{theorem}[Van Der Sluis \cite{van1969condition}]\label{thm;van}
    Let $A$ be a $n\times m$ matrix, $P$ an arbitrary diagonal $n \times n$ matrix and $E$ the row equilibrated matrix built from $A$. Then $\kappa(EA) \leq \kappa(PA)$.
\end{theorem}

For the fixed $L$ layer neural network $F$, let $F^{eq}$ denote the equilibrated network with weights $E_kW_k$, where $E_k$ is the equilibrated preconditioner corresponding to the weight matrix $W_k$. We have the following proposition.

\begin{proposition}\label{prop;row_e_reduces_weight_cond}
$\kappa(E_kW_k) \leq \kappa(W_k)$ for any $1 \leq k \leq L$. In other words the weight matrices of the network $F^{eq}$ have at least better condition number than those of the network $F$
\end{proposition}
\begin{proof}
    The matrix $W_k$ can be written as the product 
    $I_{n_{k-1}}W_k$ where 
    $I_{n_{k-1}}\in \R^{n_{k-1} \times n_{k-1}}$ is the $n_{k-1} \times n_{k-1}$ identity matrix. Applying thm. \ref{thm;van} we have 
    \begin{equation}
        \kappa(E_kW_K) \leq \kappa(I_{n_{k-1}}W_K) = 
    \kappa(W_k).
    \end{equation}
\end{proof}

For the following theorem we fix a loss function $\mathcal{L}$ see Sec. \ref{sec;Notation}. Given two neural networks $F$ and $G$ we obtain two loss functions 
$\mathcal{L}_F$ and $\mathcal{L}_G$, each depending on the weights of the respective networks. We will denote the Hessian of these two loss functions at a point $\theta$ by $H_F(\theta)$ and $H_G(\theta)$ respectively.

\begin{theorem}\label{thm;row_e_hess}
Let $F(x;\theta)$ be a fixed $L$ layer feed forward neural network. 
Let $F^{eq}(x;\theta)$ denote the equilibrated network obtained by equilibrating the weight matrices of $F$. Then 
\begin{equation}
    \kappa(H_{F^{eq}}(\theta)) \leq 
    \kappa(H_{F}(\theta))
\end{equation}
for all parameters $\theta$ at which both $H_{F^{eq}}$ and 
$H_{F}$ have full rank.
\end{theorem}

The proof of Thm. \ref{thm;row_e_hess} is given in Supp. material Sec. 1.

Thm. \ref{thm;row_e_hess} shows that weight conditioning by
equilibrating the weights of the network $F$ thereby forming $F^{eq}$ leads to a better conditioned Hessian of the loss landscape. From the discussion in sec. \ref{sec;motivation} we see that this implies that locally around points where the Hessian has full rank, a gradient descent algorithm will move faster. 

Weight conditioning can also be applied to a convolutional layer. Please see Supp. material Sec. 1 for a detailed analysis on how this is done.

%\subsection{Weight conditioning for convolutional layers}\label{sec;condition_cnn}

%Weight conditioning can also be extended to the case of convolutional architectures. Convolutional neural networks have three main types of layers, feedforward layers, convolutional layers and max pooling layers. Sec. \ref{subsec;condition_ff} shows how to weight condition a feedforward layer. We therefore focus on the case of a convolutional layer and a maxpooling layer. 

%A convolutional layer has as its building block a collection of kernels that perform the convolutional operator. Such a component is given by a 3 tensor of the form $(N, H, W)$ where $N$ denotes the number of kernels and each kernel is $H \times W$. In this setting we apply weight conditioning to each of the $N$ kernels. The right side of 
%fig. \ref{fig:equib} gives a schematic diagram of how the kernels are conditioned using an equilibrated preconditioner. For further details, and anlog of Thm. \ref{thm;row_e_hess} in this context, the reader can consult Supp. material sec. 1.

\section{Experiments}\label{sec;exps}

\subsection{Convolutional Neural Networks (CNNs)}\label{subsec;cnn}

CNNs are pivotal for vision-related tasks, thriving in image classification challenges due to their specialized architecture that efficiently learns spatial feature hierarchies. Among the notable CNN architectures in the literature, we will specifically explore two significant ones for our experiment: the Inception \cite{szegedy2015going} architecture and DenseNet \cite{huang2017densely}, both of which continue to be highly relevant and influential in the field.

%For this section we will focus on the Inception architecture and the DenseNet architecture.

\subsubsection{Experimental setup:} 
We will assess four normalization strategies on two CNN architectures. Our study compares BN, BN with weight standardization (BN + WS), BN with weight normalization (BN + W), and BN with equilibrated weight conditioning (BN + E). Training employs SGD with a 1e-3 learning rate on CIFAR10 and CIFAR100, using a batch size of 128 across 200 epochs. For more details on the training regiment, see Supp. material Sec. 2. For results on ImageNet1k Supp. material Sec. 2.

%For both architectures, we will explore the efficacy of four distinct normalization techniques. Given the depth and complexity of these models, incorporating batch normalization has proven essential for achieving stable training with Stochastic Gradient Descent (SGD). Thus, our investigation will focus on batch normalization (BN), batch normalization combined with weight standardization (BN + WS), batch normalization combined with weight normalization (BN + W), and batch normalization combined with weight conditioning (BN + E), utilizing an equilibrated conditioner as our preconditioner. Each model will be trained using SGD at a learning rate of 1e-3, and our experiments will be conducted on the CIFAR10 and CIFAR100 datasets. We used a batch size of 128 for both datasets and trained for 200 epochs. For insights into our findings on the ImageNet1k dataset, please refer to Section 2 of the Supp. material.

\subsubsection{Inception:} 
In our experiment, we employed a modified Inception architecture, as proposed in \cite{szegedy2015going}, featuring eight inception blocks. Weight conditioning was effectively applied to just the initial convolutional layer and the final linear layer. For detailed insights into the architecture and normalization applications, see Sec. 2 of the Supp. material.

\begin{figure}[t]
    \centering
    \includegraphics[width=1.0\linewidth]
    {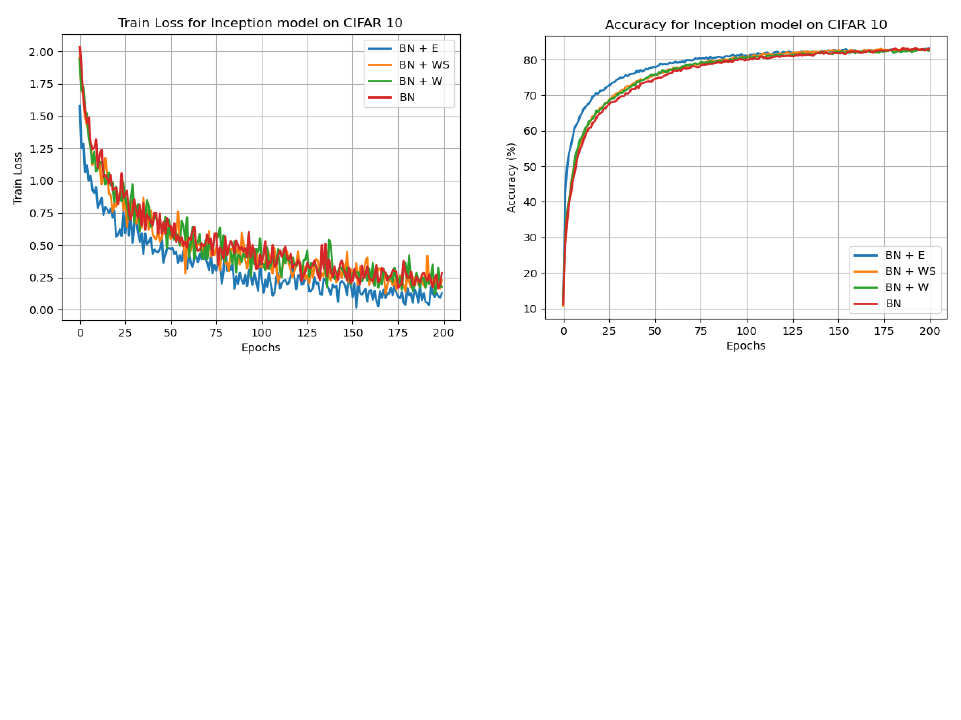}
    %\caption{Trained for 350s}
    %\end{subfigure}
    \vspace{-13em}
    \caption{Left; Train loss curves for four normalization schemes on an Inception architecture trained on the CIFAR10 dataset. Right; Top-1\% accuracy plotted during training. We see that BN + E converges the fastest.}
    \label{fig:inception_results_10}
\end{figure}

\begin{figure}[t]
    \centering
    \includegraphics[width=1.0\linewidth]
    {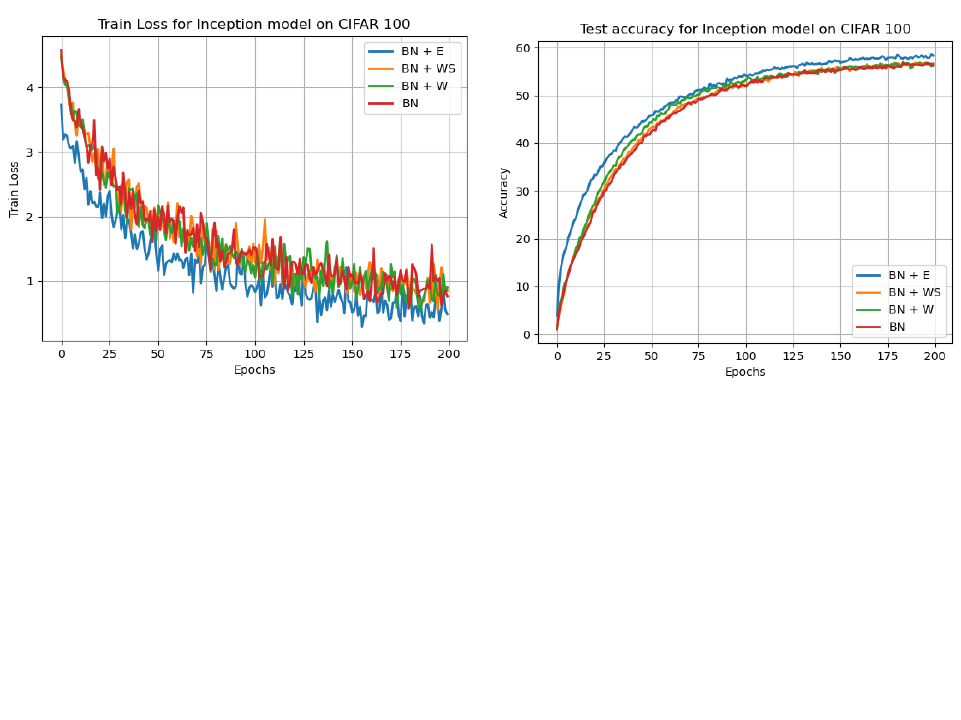}
    %\caption{Trained for 350s}
    %\end{subfigure}
    \vspace{-13em}
    \caption{Left; Train loss curves for four normalization schemes on an Inception architecture trained on the CIFAR100 dataset. Right; Top-1\% accuracy plotted during training. We see that BN + E converges the fastest.}
    \label{fig:inception_results_100}
\end{figure}

Results on the CIFAR10 dataset, depicted in Fig. \ref{fig:inception_results_10}, highlight that the Inception model with BN + E exhibits the quickest loss reduction and accelerated convergence in Top-1\% accuracy among all four normalization strategies. Similarly, Fig. \ref{fig:inception_results_100} illustrates that, on the CIFAR100 dataset, BN + E achieves the lowest training loss and highest Top-1\% accuracy, outperforming other methods in convergence speed. Across both datasets, as summarized in Tab. \ref{tab:inception_accuracy}, BN + E consistently delivers superior Top-1\% and Top-5\% accuracies, affirming its effectiveness.

%For this experiment we used the Inception architecture proposed in (reference). We used a slightly smaller architecture making use of 8 inception blocks in total. Furthermore, we found that it sufficed to apply weight conditioning on the first convolutional layer and the last linear feedforward layer.  For an in-depth description of the architecture employed in our study, and how we applied each normalization, please refer to Sec. 2 in the Supp. material.

%Fig. \ref{fig:inception_results_10} shows the results of training all 4 normalizations on the CIFAR10 dataset. As shown in the figure the train loss (left) for the Inception model with WN + E decays the fastest. The accuracy (right) gives the Top-1\% accuracy during all of training for all 4 normalizations. Although the all 4 normalizations converged to similar Top-1\% accuracies, WN + E converged faster. 

%Fig. \ref{fig:inception_results_100} shows the results for the CIFAR100 dataset. For this dataset WN + E had the lowest train loss (left of Fig. \ref{fig:inception_results_100}) and the highest Top-1\% accuracy (right of Fig. \ref{fig:inception_results_100}) at convergence. Furthermore, WN + E converges the fastest out of all 4 methods.
%Both Figs \ref{fig:inception_results_10} and \ref{fig:inception_results_100} show the Top-1\% accuracy. Tab. \ref{tab:inception_accuracy} gives the final Top-1\% and Top-5\% accuracy for all 4 normalizations on both datasets. As can be seen from that table, WN + E performs the best.

\begin{table}[ht]
\centering
\caption{Final Top-1\% and Top-5\% accuracy for the four normalizations on an Inception network trained on CIFAR10/CIFAR100.} % Caption moved above the table
\label{tab:inception_accuracy}
\begin{tabular}{|l|cc|c|cc|c|}
\hline
%\toprule
 & \multicolumn{2}{c|}{CIFAR10} & & \multicolumn{2}{c|}{CIFAR100} & \\ 
\cline{2-3} \cline{5-6}
 & Top-1\% & Top-5\% & & Top-1\% & Top-5\% & \\
%\midrule
\hline
BN + E & \textbf{83.3} & \textbf{94.6} &  & \textbf{58.7} & \textbf{71.3}& \\
BN + WS & 82.9 & 94.4 & & 56.1& 70.1 & \\
BN + W & 83.1 & 94.2 &  & 56.3&  70.4 & \\
BN & 83.1& 94.3 & &55.9 & 69.9& \\
\hline
%\bottomrule
\end{tabular}
\end{table}

\subsubsection{DenseNet:} For this experiment we tested four normalizations on the DenseNet architecture \cite{huang2017densely}. We found that it sufficed to apply equilibrated weight conditioning to the first convolutional layer and the last feedforward layer as in the case for Inception. An in-depth description of the architecture employed in our study, and how we applied each normalization is given in Sec. 2 of the Supp. material.

\begin{figure}[t]
    \centering
    \includegraphics[width=1.0\linewidth]
    {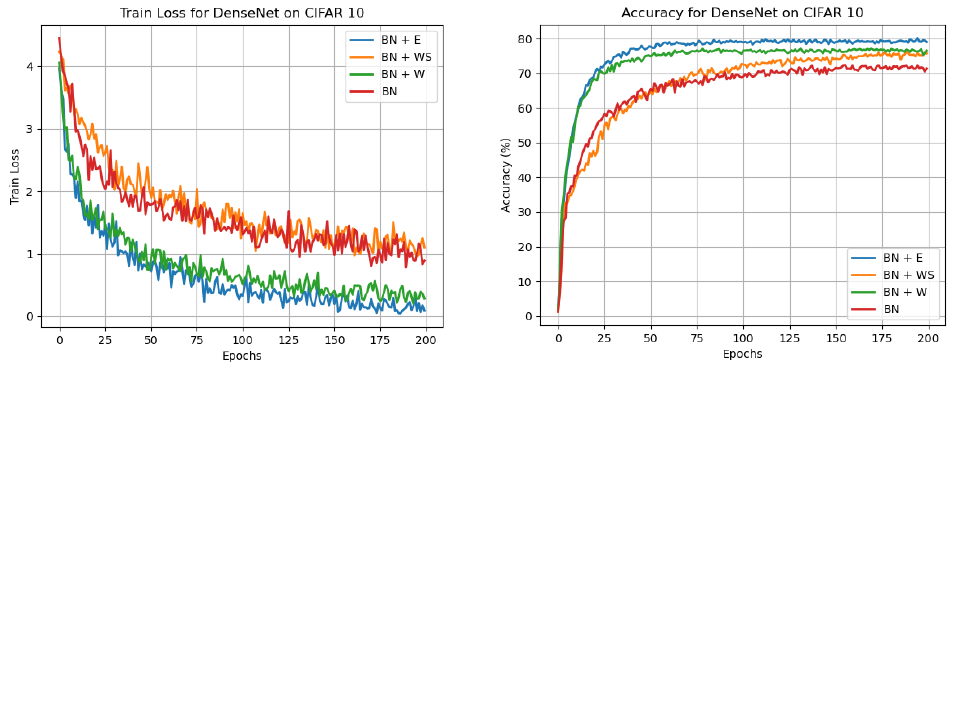}
    %\caption{Trained for 350s}
    %\end{subfigure}
    \vspace{-12em}
    \caption{Left; Train loss curves for four normalization schemes on a DenseNet architecture trained on CIFAR10. Right; Top-1\% accuracy plotted during training. We see that BN + E yields higher Top-1\% accuracy than the other three.}
    \label{fig:densenet_cifar10_results}
\end{figure}

\begin{figure}[t]
    \centering
    \includegraphics[width=1.0\linewidth]
    {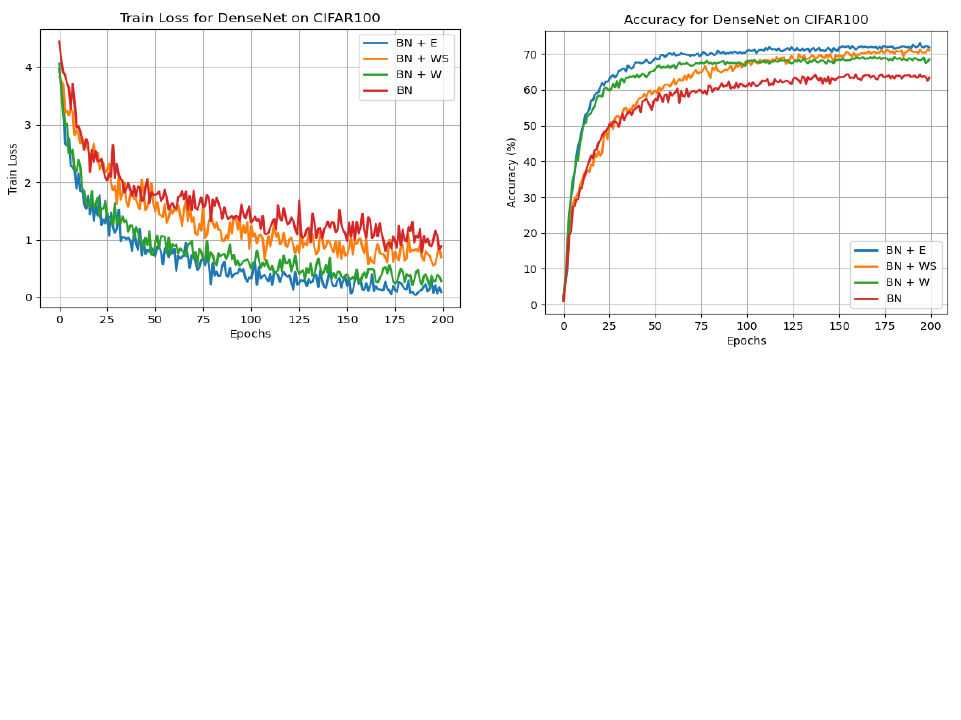}
    %\caption{Trained for 350s}
    %\end{subfigure}
    \vspace{-14em}
    \caption{Left; Train loss curves for four normalization schemes on a DenseNet architecture trained on CIFAR100. Right; Top-1\% accuracy plotted during training. We see that BN + E and BN + W converge much faster than the other two, with BN + E reaching a higher accuracy.}
    \label{fig:densenet_results_cifar100}
\end{figure}

Fig. \ref{fig:densenet_cifar10_results} and Fig. \ref{fig:densenet_results_cifar100} display CIFAR10 and CIFAR100 dataset results, respectively. For CIFAR10, BN + E demonstrates quicker convergence and higher Top-1\% accuracy as shown in the train loss curves and accuracy figures. CIFAR100 results indicate BN + E outperforms other normalizations in both convergence speed and Top-1\% accuracy. Tab. \ref{tab:densenet_accuracy} summarizes the final Top-1\% and Top-5\% accuracies for all normalizations, with BN + E leading in performance across both datasets.

%Fig. \ref{fig:densenet_cifar10_results} shows the results on the CIFAR10 dataset. The left of the figure shows the train loss curves showing that both BN + E and BN + W perform well with BN + E converging faster. The right of the figure shows the Top-1\% accuracy during training. Both WN + E and WN + W converge the fastest with WN + E reaching a higher accuracy.

%Fig. \ref{fig:densenet_results_cifar100} shows the results on the CIFAR100 dataset. The left of the figure shows the train loss of all 4 normalizations, while BN + WS, BN + W and BN all perform similary BN + E reaches convergence at a faster rate. On the right of the figure, it can be seen that BN + E converges to a higher Top-1\% accuracy at a faster rate.

%Tab. \ref{tab:densenet_accuracy} shows the final Top-1\% and Top-5\% accuracy values for all 4 normalizations on both the CIFAR10 and CIFAR100 datasets. As can be seen from that table WN + E performs the best.

\begin{table}[ht]
\centering
\caption{Final Top-1\% and Top-5\% accuracy for the four normalizations on a DenseNet network trained on CIFAR10/CIFAR100} % Caption moved above the table
\label{tab:densenet_accuracy}
\begin{tabular}{|l|cc|c|cc|c|}
\hline
%\toprule
 & \multicolumn{2}{c|}{CIFAR10} & & \multicolumn{2}{c|}{CIFAR100} & \\ 
\cline{2-3} \cline{5-6}
 & Top-1\% & Top-5\% & & Top-1\% & Top-5\% & \\
%\midrule
\hline
BN + E & \textbf{79.96} & \textbf{92.1} &  & \textbf{72.4} & \textbf{83.3}& \\
BN + WS & 75.9 & 90.2 & & 71.7& 82.9 & \\
BN + W & 76.2 & 90.9 &  & 68.8&  83.1 & \\
BN & 70.5& 89.5 & &64.6 & 80.4& \\
\hline
%\bottomrule
\end{tabular}
\end{table}

\subsection{Vision Transformers (ViTs)}

Vision Transformers (ViTs) \cite{dosovitskiy2020image} have emerged as innovative architectures in computer vision, showcasing exceptional capabilities across a broad spectrum of tasks. In general, we found that most work in the literature applied layer normalization to vision transformers which was much more robust for training than batch normalization. We found that if we removed layer normalization training was impeded significantly leading to gradients not being able to be backpropgated. Therefore, for this experiment we will consider 3 normalization scenarios, layer normalization (LN), layer normalization with weight normalization (LN + W) and layer normalization with equilibrated weight conditioning (LN + E).

%\subsubsection{ViT-small (ViT-S):} For the first experiment, we will train a small Vit on the CIFAR100 dataset. We found that it sufficed to apply weight conditioning on the heads and the feedforward part of the transformer block. The same was done with weight conditioning. Layer normalization was applied as in (reference). For a detailed account of the architecture and how each normalization was applied please refer to Sec. 2 of the Supp. material.

%Fig. \ref{fig:vits_cifar100} shows the results of applying the above 3 normalizations and training from scratch on the CIFAR100 dataset. As can be seen from that figure, LN + E is superior over the other 2 normalizations. Tab. \ref{tab:vits_acc} shows the final Top-1\% and Top-5\% accuracy. As can be seen from the table, LN + E performs the best. We note that the accuracies obtained might seem rather low, this is because ViT-S has 22 million parameters which is generally too much for dataset such as CIFAR100.

%\begin{figure}[t]
%    \centering
%    \includegraphics[width=1.0\linewidth]
%    {figs/exps/vit/vit_c100}
%    %\caption{Trained for 350s}
%    %\end{subfigure}
%    \vspace{-15em}
%    \caption{sad}
%    \label{fig:vits_cifar100}
%\end{figure}

%\begin{table}[ht]
%\centering
%\caption{Your caption here.}
%\label{tab:vits_acc}
%\begin{tabular}{|l|cc|}
%\hline
%\multicolumn{1}{|c|}{} & \multicolumn{2}{c|}{CIFAR100} \\
%\cline{2-3}
%\multicolumn{1}{|c|}{} & Top-1\% & Top-5\% \\
%\hline
%BN + E & \textbf{55.3} & \textbf{67.8} \\
%BN + W & 50.1 & 64.4 \\
%BN & 49.9 & 64.1 \\
%\hline
%\end{tabular}
%\end{table}

\subsubsection{ViT-Base (ViT-B):} The ViT-B architecture \cite{steiner2021train}, with its 86 million parameters, exemplifies a highly overparameterized neural network. We investigated three variations of ViT-B, each modified with a different normalization: LN, LN + W, and LN + E; the implementation details are provided in Sec. 2 of the Supp. material. These models were trained on the ImageNet1k dataset using a batch size of 1024 and optimized with AdamW.

Tab. \ref{tab:vitbase_acc} shows the final Top-1\% and Top-5\% accuracies for the ViT-B architecture with the above three different normalizations. The table shows that LN + E outperformed the other normalization schemes.

%Fig. \ref{fig:vitb_imagenet} presents the training loss and Top-1\% accuracy across these variations, highlighting a narrower performance disparity between LN + E and the others compared to results on datasets like CIFAR100. Tab. \ref{tab:vitbase_acc} summarizes their final Top-1\% and Top-5\% accuracies.

%The ViT-B architecture is highly overparameterized neural architecture with 86 million parameters in total \cite{steiner2021train}. We trained 3 ViT-B architectures applying on each one the normalizations LN, LN + W, and LN + E. For details on how these normalizations were applied, see Sec. 2 of the Supp. material. Each of the ViT-B's were trained on the ImageNet 1k dataset with a batch size of 1024. All three architectures were trained with the AdamW optimizer.

%Fig. \ref{fig:vitb_imagenet} shows the results of the train loss and 
%Top-1\% accuracy for the three transformers. In this case we found that the gap between LN + E and the other two was much smaller than on datasets like CIFAR100. Tab. \ref{tab:vitbase_acc} shows the final Top-1\% and Top-5\% accuracies for the three transformers. 

\subsubsection{ViT-Small (ViT-S):} For experiments on smaller transformer networks on the CIFAR100 dataset, please see Sup. material sec. 2.

%\begin{figure}[t]
%    \centering
%    \includegraphics[width=0.9\linewidth]
%    {figs/exps/vit/vit_image}
    %\caption{Trained for 350s}
    %\end{subfigure}
%    \vspace{-12em}
%    \caption{Left; Train loss for the three normalizations on a ViT-B architecture on ImageNet1k. Right; Top-1\% accuracy during training. We see in this case that LN+E obtains a slightly better accuracy.}
%    \label{fig:vitb_imagenet}
%\end{figure}

\begin{table}[ht]
\centering
\caption{Final Top-1\% and Top-5\% accuracy for the three normalizations on a ViT-B architecture on ImageNet1k.}
\label{tab:vitbase_acc}
\begin{tabular}{|l|cc|}
\hline
\multicolumn{1}{|c|}{} & \multicolumn{2}{c|}{ImageNet1k} \\
\cline{2-3}
\multicolumn{1}{|c|}{} & Top-1\% & Top-5\% \\
\hline
LN + E & \textbf{80.2} & \textbf{94.6} \\
LN + W & 80.0 & 94.4 \\
LN & 79.9 & 94.3 \\
\hline
\end{tabular}
\end{table}

\subsection{Neural Radiance Fields (NeRF)}

\begin{figure}[t]
    \centering
    \includegraphics[width=0.8\linewidth]
    {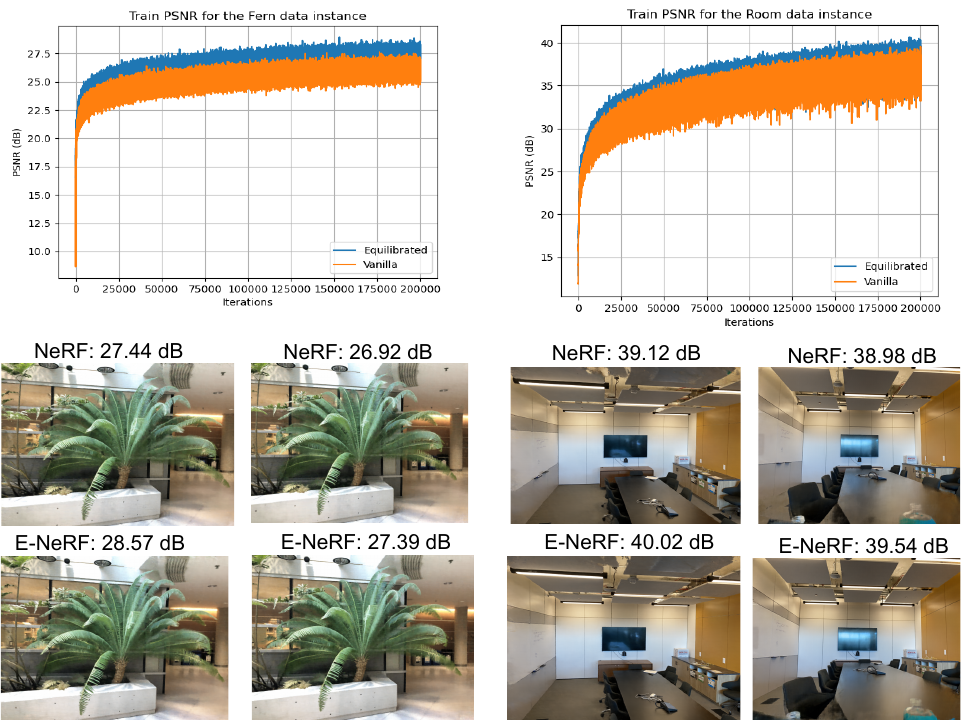}
    %\caption{Trained for 350s}
    %\end{subfigure}
    %\vspace{-1em}
    \caption{Top; Train PSNR curves for NeRF and E-NeRF on the Fern instance (left) and Room instance (right) from the LLFF dataset. Bottom; Comparison of NeRF and E-NeRF on two test scenes for the Fern instance (left) and Room instance (right). In each case E-NeRF has superior performance (zoom in for better viewing).}
    \label{fig:nerf_results_llff}
\end{figure}

Neural Radiance Fields (NeRF) \cite{nerf, chng2022gaussian, lin2021barf, reiser2021kilonerf} have emerged as a pioneering approach in 3D modeling, using Multi-Layer Perceptrons (MLPs) to reconstruct 3D objects and scenes from multi-view 2D images. We utilized the standard NeRF model from \cite{nerf}, noting that unlike transformers or CNNs, NeRF architectures are relatively shallow, typically comprising 8 hidden layers, where common normalization techniques can hinder training. We explored the performance of a standard NeRF against an equilibrated weight conditioned NeRF (E-NeRF). For an in-depth look at the NeRF setup and our application of weight conditioning, refer to Sec. 2 in the Supp. material. Both models were trained on the LLFF dataset \cite{nerf}.

Fig. \ref{fig:nerf_results_llff} presents outcomes for the Fern and Room instances from LLFF, demonstrating that E-NeRF outperforms the vanilla NeRF by an average of 0.5-1 dB. Tab. \ref{tab:nerf_results} gives the test PSNR averaged over three unseen scenes over the whole LLFF dataset. E-NeRF on average performs better over the whole dataset.

\begin{table}[ht]
\centering
\caption{Final test PSNRs for NeRF and E-NeRF on the LLFF dataset averaged over all three test scenes.}
\label{tab:nerf_results}
\begin{tabular}{lccccccccc}
\toprule
 & \multicolumn{9}{c}{PSNR (dB) $\uparrow$} \\
\cmidrule{2-10}
 & Fern & Flower & Fortress & Horns & Leaves & Orchids & Room & Trex & Avg. \\
\midrule
E-NeRF & 28.53 & 31.8 & 33.14 & 29.62 & 23.84 & 24.14 & 39.89 & 30.67 & \textbf{30. 20}\\
NeRF   & 27.51 & 31.3 & 33.16 & 29.34  & 23.10 & 23.98 & 39.15 & 30.05 & 29.6\\
\bottomrule
\end{tabular}
\end{table}

%\subsection{3D Shape Modelling}

\subsection{Further Experiments}\label{subsec;further_exps}

Further experiments can be found in Supp. material Sec. 3:
Applications to 3D shape modelling, cost analysis and ablations.

\section{Limitations}

Weight conditioning entails the application of a preconditioner to the weight matrices of a neural network during the forward pass, which does extend the training duration per iteration of a gradient optimizer compared to other normalization methods. We leave it as future work to develop methods to bring down this cost.
%This opens up an intriguing avenue for future research focused on developing cost-effective preconditioning strategies that optimize training efficiency.

%, offers a trade-off. Despite increasing iteration times, weight conditioning typically leads to faster convergence with fewer steps required. 
%This opens up an intriguing avenue for future research focused on developing cost-effective preconditioning strategies that optimize training efficiency.

\section{Conclusion}
In this work, we introduced weight conditioning to improve neural network training by conditioning weight matrices, thereby enhancing optimizer convergence. We developed a theoretical framework showing that weight conditioning reduces the Hessian's condition number, improving loss function optimization. Through empirical evaluations on diverse deep learning models, we validated our approach, confirming that equilibrated weight conditioning consistently aligns with our theoretical insights.

%In this work, we presented weight conditioning, a technique designed to condition the weight matrices of neural networks, enhancing the convergence of gradient-based optimizers. We established a theoretical framework demonstrating how weight conditioning effectively reduces the condition number of the Hessian associated with a neural network's loss function. Further, we proposed equilibrated weight conditioning as an effective strategy for this purpose and conducted empirical evaluations across a wide variety of deep learning models, confirming our theoretical findings in every instance.

%\clearpage  % TODO REVIEW/FINAL: This \clearpage needs to be removed from both review and camera-ready versions.

\section*{Acknowledgements}
% Please insert your acknowledgments here.
Simon Lucey acknowledges support from the Australian Research Council (ARC) through the Discovery Project DP220103803.
Thomas X Wang acknowledges financial support provided by DL4CLIM (ANR-
19-CHIA-0018-01), DEEPNUM (ANR-21-CE23-0017-02), PHLUSIM (ANR-23-CE23-
0025-02), and PEPR Sharp (ANR-23-PEIA-0008”, “ANR”, “FRANCE 2030”). 
Part of this work was performed using HPC resources from GENCI–IDRIS (Grant 2023-AD011013332R1).

\bibliographystyle{splncs04}
\bibliography{main}
\end{document}